\documentclass[11pt,letterpaper]{article}
\usepackage{emnlp2017}
\usepackage{times}
\usepackage{latexsym}

\usepackage{url}
\usepackage{CJK}
\usepackage{multicol}
\usepackage{multirow}
\usepackage{array}
\usepackage{amsmath}
\usepackage{balance}
\usepackage{graphicx}
\usepackage{enumitem}% http://ctan.org/pkg/enumitem
\emnlpfinalcopy

\def\emnlppaperid{716}

\title{Exploiting Cross-Sentence Context for Neural Machine Translation}
\def\fndaff{$^\dagger$}
\def\sstaff{$^\ddag$}
\author{Longyue Wang\fndaff ~~~~ Zhaopeng Tu\sstaff\thanks{~~Corresponding Author: Zhaopeng Tu} ~~~~ Andy Way\fndaff ~~~~ Qun Liu\fndaff\\
\fndaff ADAPT Centre, School of Computing, Dublin City University, Ireland \\ 
{\tt \{longyue.wang, andy.way, qun.liu\}@adaptcentre.ie}\\
\sstaff Tencent AI Lab, China\\
{\tt tuzhaopeng@gmail.com}}

\date{}

\begin{document}

\maketitle

\begin{abstract}
   In translation, considering the document as a whole can help to resolve ambiguities and inconsistencies. In this paper, we propose a cross-sentence context-aware approach and investigate the influence of historical contextual information on the performance of neural machine translation (NMT). First, this history is summarized in a hierarchical way. We then integrate the historical representation into NMT in two strategies: 1) a warm-start of encoder and decoder states, and 2) an auxiliary context source for updating decoder states. Experimental results on a large Chinese-English translation task show that our approach significantly improves upon a strong attention-based NMT system by up to +2.1 BLEU points.
   
%   In translation, considering the document as a whole allows certain ambiguities and inconsistencies to be resolved.
  %  We propose context-based NMT models, which can summarize the contextual sentences in a hierarchical way. Depending on how to integrate the different historic annotations into NMT, we build various context-aware translation models. Finally, we compare and analyze the performance of different variants on a large Chinese-English corpus. Experimental results show that our approach significantly improves upon a standard attention-based NMT system by +2.7 BLEU points at most.
  %As source history and target history have a direct impact on the adequacy and fluency of a translation, we explore both source and target history as well as encoder and decoder. Depending on how we integrate the historic annotations into decoder/encoder, our various context-based models are classified into two broad categories, initial and static. Finally, we compare and analyze all the variants of our context-based models on a large corpus. Experimental results show that our approach significantly improves upon a standard attention-based NMT system by +1.14 BLEU points.
 
\end{abstract}

\section{Introduction}
\label{sec:1}

Neural machine translation (NMT) has been rapidly developed in recent years \cite{D13-1176,sutskever2014sequence,bahdanau2015neural,Tu:2016:ACL}. The encoder-decoder architecture is widely employed, in which the encoder summarizes the source sentence into a vector representation, and the decoder generates the target sentence word by word from the vector representation. Using the encoder-decoder framework as well as gating and attention techniques, it has been shown that the performance of NMT has surpassed the performance of traditional statistical machine translation (SMT) on various language pairs \cite{Luong2015}.

The continuous vector representation of a symbol encodes multiple dimensions of similarity, equivalent to encoding more than one meaning of a word.
Consequently, NMT needs to spend a substantial amount of its capacity in disambiguating source and target words based on the context defined by a source sentence \cite{choi2016context}. Consistency is another critical issue in document-level translation, where a repeated term should keep the same translation throughout the whole document \cite{xiao2011document, carpuat2012trouble}. 
Nevertheless, current NMT models still process a documents by translating each sentence alone, suffering from inconsistency and ambiguity arising from a single source sentence. These problems are difficult to alleviate using only limited intra-sentence context.

%isolated sentences, disregarding significant discourse information beyond sentence level. It is difficult to deal with ambiguity or inconsistency problems using such limited intra-sentential context.}

The cross-sentence context, or global context, has proven helpful to better capture the meaning or intention in sequential tasks such as query suggestion \cite{Sordoni2015A} and dialogue modeling \cite{vinyals2015neural,Serban:2016}.
The leverage of global context for NMT, however, has received relatively little attention from the research community.\footnote{To the best of our knowledge, our work and~\newcite{jean2017does} are two independently early attempts to model cross-sentence context for NMT.}
In this paper, we propose a cross-sentence context-aware NMT model, which considers the influence of {\em previous source sentences} in the same document.\footnote{In our preliminary experiments, considering target-side history inversely harms translation performance, since it suffers from serious error propagation problems.} %One possible reason is that, while the target-side history is summarized from ground-truth references in training, it is from previous generated translations, which can be erroneous, at test time.}

Specifically, we employ a hierarchy of Recurrent Neural Networks (RNNs) to summarize the cross-sentence context from source-side previous sentences, which deploys an additional document-level RNN on top of the sentence-level RNN encoder~\cite{Sordoni2015A}. After obtaining the global context, we design several strategies to integrate it into NMT to translate the current sentence:

\begin{itemize}
    \setlength\itemsep{-0.1em}
    \item {\em Initialization}, that uses the history representation as the initial state of the encoder, decoder, or both;
    \item {\em Auxiliary Context}, that uses the history representation as static cross-sentence context, which works together with the dynamic intra-sentence context produced by an attention model, to good effect.
    \item {\em Gating Auxiliary Context}, that adds a gate to Auxiliary Context, which decides the amount of global context used in generating the next target word at each step of decoding.
\end{itemize}

%\citet{Tu:2016:TACL} reveal that source- and target-side intra-sentence contexts have a direct impact on NMT adequacy and fluency, respectively. 

%(coherence and cohesion) between sentences in document.
%Our contributions are two-fold:
%\begin{itemize}
%    \item[1.] We extend the HRED framework to better suit the translation task. After summarizing the history by means of a hierarchical RNN, we propose two strategies to integrate it into NMT: 1) {\em Initialization}, that uses the history representation as the initial state of the encoder, decoder, or both; and 2) {\em Auxiliary Context}, that uses the history representation as static inter-sentence context, which works together with the dynamic intra-sentence context produced by an attention model, to good effect.
%    \item [2.] We investigate the influence of source- and target-side history on the performance of NMT. In this work, however, we find that source-side history is more useful than target-side context, since the latter suffers from serious error propagation problems.
%To condition the prediction of the next sentence on the previous sentences in the document, we deploy an additional, document-level RNN on top of the sentence-level RNN encoder, thus forming a hierarchy of RNNs.
%\end{itemize}
% Second, we investigate the influence of different factors on the performance of NMT.  Thus, we explore different sides of context information (i.e. source, target and both) to the translation performance. Besides, we also investigate the ways of incorporating the context information into NMT (i.e. static, initial and both).

Experimental results show that the proposed {\em initialization} and {\em auxiliary context} (w/ or w/o gating) mechanisms significantly improve translation performance individually, and combining them achieves further improvement. 
%Some interesting findings are that 1, source context can improve the translation quality while target information brings noise to translation; 2, static integration methods are more powerful than initial ones; 3, combining two best models can substantially improve the performance.

%The rest of the paper is organized as follows. In Section 2, we describe related work on DP translation. In Section 3, we present our approaches to build DP corpus, DP generator and SMT integration. Section 4 reports the experimental results of both DP generator and translation followed by a conclusion in Section 5.

%demonstrate that it is superior to state-of-the-art NMT and SMT in the literature. We extend the model architecture to better suit the translation task. 
%We focus on how to capture sentence discourse and sentence by means of such embeddings.

%and demonstrate that this model is superior to state-of-the-art NMT and SMT.

%In this work, we propose to build a context-aware NMT to improve the document-level translation. We define the problem as modeling the sentences and discourse of the document.

%We empirically investigate our hypothesis: whether source and target history correlate to translation adequacy and fluency?

% \input{figures/context_hier.tex}

\section{Approach}

%Suppose that a document consists of $M$ sentences $\{\mathbf{x}_1,\dots, \mathbf{x}_M\}$. 
Given a source sentence $\mathbf{x}_m$ to be translated, we consider its $K$ previous sentences in the same document as cross-sentence context $C=\{\mathbf{x}_{m-K},...,\mathbf{x}_{m-1}\}$. In this section, we first model $C$, which is then integrated into NMT. 

\begin{figure}[h]
\centering
\graphicspath{ {figures/} }
\includegraphics[width=0.46\textwidth]{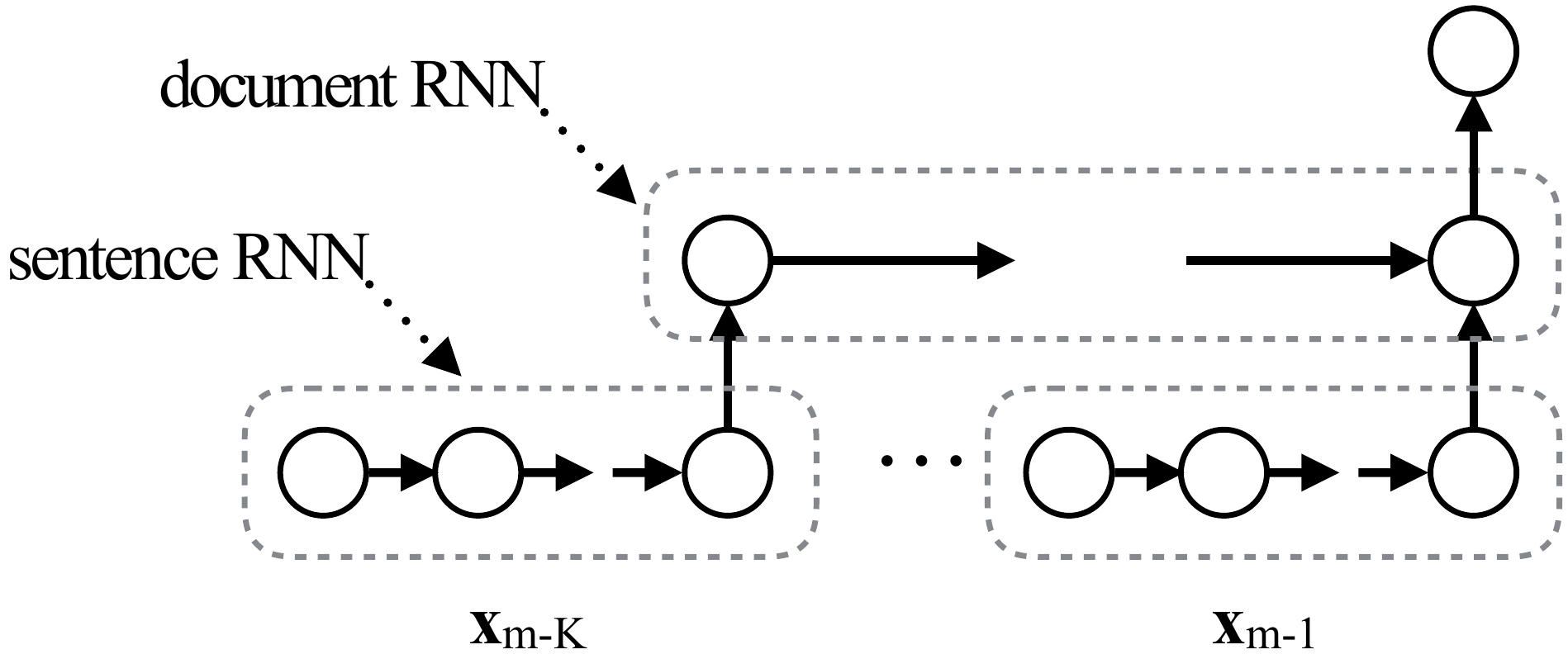}
\caption{Summarizing global context with a hierarchical RNN (${\bf x}_k$ is the $k$-th source sentence).}
\label{fig:1}
\end{figure}

\subsection{Summarizing Global Context}
\label{sec:2.1}

As shown in Figure \ref{fig:1}, we summarize the representation of $C$ in a hierarchical way:

\paragraph{Sentence RNN}
For a sentence $\mathbf{x}_k$ in $C$, the sentence RNN reads the corresponding words $\{x_{1,k}, ..., x_{n,k}, \dots, x_{N,k}\}$ sequentially and updates its hidden state:
\begin{equation}
h_{n,k} = f(h_{n-1,k},x_{n,k})
\end{equation}
where $f(\cdot)$ is an activation function, and $h_{n,k}$ is the hidden state at time $n$. The last state $h_{N,k}$ stores order-sensitive information about all the words in $\mathbf{x}_k$, which is used to represent the summary of the whole sentence, i.e. ${S}_k\equiv h_{N,k}$. After processing each sentence in $C$, we can obtain all sentence-level representations, which will be fed into document RNN.

\paragraph{Document RNN} It takes as input the sequence of the above sentence-level representations $\{{S}_{1}, ..., {S}_{k},...,{S}_{K}\}$ and computes the hidden state as:
\begin{equation}
h_k = f(h_{k-1}, {S}_k)
\end{equation}
where $h_k$ is the recurrent state at time $k$, which summarizes the previous sentences that have been processed to the position $k$. Similarly, we use the last hidden state to represent the summary of the global context, i.e. $D \equiv h_{K}$.
% such as a generalization or specification

%As discussed in Section \ref{sec:1}, $hist$ can be summarized from source-, target- or two-side context. In our experiments, we explore how historical information on different sides affects translation performance.   

% (i.e. $hist_{src}$, $hist_{tgt}$, $hist_{both}$)

% The context can be source side $C_{src}=\{x_{m-1+N},...,x_{m-1}\}$ or target side $C_{tgt}=\{y_{m-1+N},...,y_{m-1}\}$ or both. 

\subsection{Integrating Global Context into NMT}
\label{sec:2.2}

\begin{figure*}[t]
\centering
\graphicspath{ {figures/} }
\includegraphics[width=0.75\textwidth]{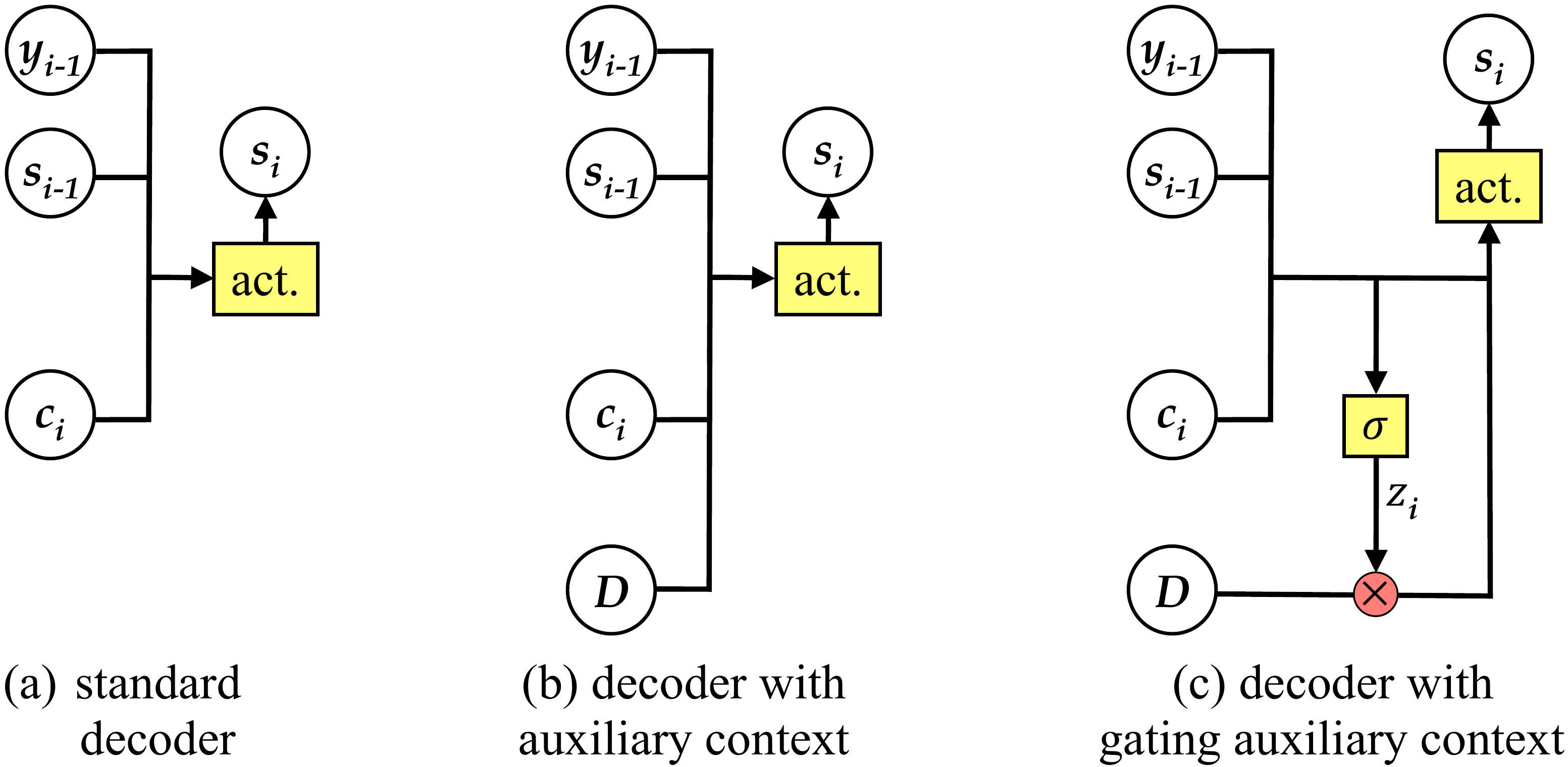}
\caption{Architectures of NMT with auxiliary context integrations. $act.$ is the decoder activation function, and $\sigma$ is a sigmoid function.}
\label{fig:2}
\end{figure*}

We propose three strategies to integrate the history representation $D$ into NMT:

\paragraph{Initialization} We use $D$ to initialize either NMT encoder, NMT decoder or both. For encoder, we use $D$ as the initialization state rather than all-zero states as in the standard NMT~\cite{bahdanau2015neural}. 
For decoder, we rewrite the calculation of the initial hidden state $s_0 = \tanh(W_s h_N)$ as 
$s_0 = \tanh(W_s h_N + W_D D)$
where $h_N$ is the last hidden state in encoder and $\{W_s, W_D\}$ are the corresponding weight metrices.

\paragraph{Auxiliary Context}
In standard NMT, as shown in Figure~\ref{fig:2} (a), the decoder hidden state for time $i$ is computed by
\begin{equation}
s_i = f(s_{i-1}, y_{i-1}, c_i)
\end{equation}
where $y_{i-1}$ is the most recently generated target word, and $c_i$ is the intra-sentence context summarized by NMT encoder for time $i$. As shown in Figure~\ref{fig:2} (b), {\em Auxiliary Context} method adds the representation of cross-sentence context $D$ to jointly update the decoding state $s_i$:
\begin{equation}
s_i=f(s_{i-1}, y_{i-1}, c_i, D)
\end{equation}
In this strategy, $D$ serves as an auxiliary information source to better capture the meaning of the source sentence. Now the gated NMT decoder has four inputs rather than the original three ones. The concatenation $[c_i, D]$, which embeds both intra- and cross-sentence contexts, can be fed to the decoder as a single representation. We only need to modify the size of the corresponding parameter matrix for least modification effort.

\paragraph{Gating Auxiliary Context} 
The starting point for this strategy is an observation: the need for information from the global context differs from step to step during generation of the target words. For example, global context is more in demand when generating target words for ambiguous source words, while less by others. To this end, we extend auxiliary context strategy by introducing a context gate~\cite{Tu:2017:TACL} to dynamically control the amount of information flowing from the auxiliary global context at each decoding step, as shown in Figure~\ref{fig:2} (c).

% generating target words at different steps have the different needs of the global context. 

Intuitively, at each decoding step $i$, the context gate looks at decoding environment (i.e., $s_i$, $y_{i-1}$, and $c_i$), and outputs a number between 0 and 1 for each element in $D$, where 1 denotes ``completely transferring this'' while 0 denotes ``completely ignoring this''. The global context vector $D$ is then processed with an element-wise multiplication before being fed to the decoder activation layer.

Formally, the context gate consists of a sigmoid neural network layer and an element-wise multiplication operation. It assigns an element-wise weight to $D$, computed by
\begin{equation}
z_i =  \sigma (U_z s_{i-1} + W_z y_{i-1} + C_z c_i)
\label{eqn-context-gate}
\end{equation}
Here $\sigma(\cdot)$ is a logistic sigmoid function, and $\{W_z, U_z, C_z\}$ are the weight matrices, which are trained to learn when to exploit global context to maximize the overall translation performance.
Note that $z_i$ has the same dimensionality as $D$, and thus each element in the global context vector has its own weight.
Accordingly, the decoder hidden state is updated by
\begin{equation}
s_i=f(s_{i-1}, y_{i-1}, c_i, z_i \otimes D)
\end{equation}

\section{Experiments}
\subsection{Setup}

We carried out experiments on Chinese--English translation task. As the document information is necessary when selecting the previous sentences, we collect all LDC corpora that contain document boundary. The training corpus consists of 1M sentence pairs extracted from LDC corpora\footnote{The LDC corpora indexes are: 2003E07, 2003E14, 2004T07, 2005E83, 2005T06, 2006E24, 2006E34, 2006E85, 2006E92, 2007E87, 2007E101, 2007T09, 2008E40, 2008E56, 2009E16, 2009E95.} with 25.4M Chinese words and 32.4M English words.
We chose the NIST05 (MT05) as our development set, and NIST06 (MT06) and NIST08 (MT08) as test sets. 
We used case-insensitive BLEU score \cite{Papineni:2002} as our evaluation metric, and sign-test \cite{Collins05} for calculating statistical significance. 

We implemented our approach on top of an open source attention-based NMT model, Nematus\footnote{Available at https://github.com/EdinburghNLP/nematus.} \cite{W16-2209,sennrich2017nematus}.
We limited the source and target vocabularies to the most frequent 35K words in Chinese and English, covering approximately 97.1\% and 99.4\% of the data in the two languages respectively.
We trained each model on sentences of length up to 80 words in the training data with early stopping. 
The word embedding dimension was 600, the hidden layer size was 1000, and the batch size was 80.
%We trained our models until the BLEU score on the development set stops improving.
All our models considered the previous three sentences (i.e., $K=3$) as cross-sentence context. % {\color{red}We use early stopping.}

\begin{table*}[h]
	\begin{center}
		\begin{tabular}{|c|l|l|lll|c|}
			\hline 
			\bf \# & \bf System & \bf MT05 & \bf MT06 & \bf MT08 & \bf Ave. & {\bf $\bigtriangleup$}  \\ \hline
			1  &  \textsc{Moses} & 33.08 & 32.69 & 23.78 & 28.24   &   --\\
			\hline \hline
			2  &  \textsc{Nematus} & 34.35 & 35.75 & 25.39 & 30.57   &   --\\ \hline
			3  & +Init\textsubscript{enc} & 36.05 & 36.44\textsuperscript{$\dagger$} & 26.65\textsuperscript{$\dagger$} & 31.55 & +0.98\\
			4  & +Init\textsubscript{dec} & 36.27 & 36.69\textsuperscript{$\dagger$} & 27.11\textsuperscript{$\dagger$} & 31.90 & +1.33\\
			5  & +Init\textsubscript{enc+dec} & 36.34 & 36.82\textsuperscript{$\dagger$} & 27.18\textsuperscript{$\dagger$} & 32.00   &   +1.43\\
			\hline
			6  &  +Auxi & 35.26 & 36.47\textsuperscript{$\dagger$} & 26.12\textsuperscript{$\dagger$} & 31.30  &   +0.73\\
			7  &  +Gating Auxi & 36.64 & 37.63\textsuperscript{$\dagger$} & 26.85\textsuperscript{$\dagger$} & 32.24 & +1.67\\
			\hline
			8  &  +Init\textsubscript{enc+dec}+Gating Auxi & {\bf 36.89} & {\bf 37.76}\textsuperscript{$\dagger$} & {\bf 27.57}\textsuperscript{$\dagger$} & {\bf 32.67} &+2.10\\
			\hline
		\end{tabular}
	\end{center}
	\caption{\label{result-table} Evaluation of translation quality.  ``Init'' denotes Initialization of encoder (``enc''), decoder (``dec''), or both (``enc+dec''), and ``Auxi'' denotes Auxiliary Context. ``$\dagger$'' indicates statistically significant difference ($P < 0.01$) from the baseline \textsc{Nematus}.}
\end{table*}

\subsection{Results}
\label{sec:3.2}

Table \ref{result-table} shows the translation performance in terms of BLEU score. Clearly, the proposed approaches significantly outperforms baseline in all cases.

\paragraph{Baseline} (Rows 1-2)
\textsc{Nematus} significantly outperforms Moses -- a commonly used phrase-based SMT system \cite{Koehn:2007}, by 2.3 BLEU points on average, indicating that it is a strong NMT baseline system. It is consistent with the results in~\cite{Tu:2017:AAAI} (i.e., 26.93 vs. 29.41) on training corpora of similar scale.

\paragraph{Initialization Strategy} (Rows 3-5)
Init\textsubscript{enc} and Init\textsubscript{dec} improve translation performance by around +1.0 and +1.3 BLEU points individually, proving the effectiveness of warm-start with cross-sentence context. Combining them achieves a further improvement. 

\paragraph{Auxiliary Context Strategies} (Rows 6-7)
The gating auxiliary context strategy achieves a significant improvement of around +1.0 BLEU point over its non-gating counterpart. This shows that, by acting as a critic, the introduced context gate learns to distinguish the different needs of the global context for generating target words.

\paragraph{Combining} (Row 8)
Finally, we combine the best variants from the initialization and auxiliary context strategies, and achieve the best performance, improving upon \textsc{Nematus} by +2.1 BLEU points. This indicates the two types of strategies are complementary to each other.

\subsection{Analysis}

We first investigate to what extent the mis-translated errors are fixed by the proposed system. We randomly select 15 documents (about 60 sentences) from the test sets. As shown in Table \ref{error-table}, we count how many related errors: i) are made by NMT ({\em Total}), and ii) fixed by our method ({\em Fixed}); as well as iii) newly generated ({\em New}). About {\em Ambiguity}, while we found that 38 words/phrases were translated into incorrect equivalents, 76\% of them are corrected by our model. Similarly, we solved 75\% of the {\em Inconsistency} errors including lexical, tense and definiteness (definite or indefinite articles) cases. However, we also observe that our system brings relative 21\% new errors. 

\begin{table}[h]
\centering
\begin{tabular}{|c|c|c|c|}
\hline 
\bf Errors & \bf Ambiguity & \bf Inconsistency & \bf All\\ \hline
\bf Total  & 38 & 32 & 70 \\ \hline
\bf Fixed & 29 & 24 & 53  \\\hline
\bf New & 7 & 8 & 15 \\ \hline
\end{tabular}
\caption{\label{error-table} Translation error statistics.}
\end{table}

\begin{CJK}{UTF8}{gbsn}
\begin{table}[h]
\renewcommand\arraystretch{1.2}
\centering
\begin{tabular}{|c|m{5.7cm}|}
\hline 
Hist. & 这 不 等于 明着 提前 告诉 贪官 们 赶紧 转移 罪证 吗 ?\\ \hline
Input & 能否 遏制 和 震慑 腐官 ?\\ \hline
Ref. & Can it inhibit and deter corrupt officials?\\ \hline
NMT & Can we contain and deter the {\em \color{blue} enemy}?\\ \hline
Our & Can it contain and deter the {\bf \color{red} corrupt officials}?\\\hline
\end{tabular}
\caption{Example translations. We italicize some {\em \color{blue} mis-translated} errors and highlight the {\bf \color{red} correct} ones in bold.}
\end{table}

% \begin{CJK}{UTF8}{gbsn}

\paragraph{Case Study} Table 3 shows an example. The word ``腐官'' ({\em corrupt officials}) is mis-translated as ``{\em enemy}'' by the baseline system. With the help of the similar word ``贪官'' in the previous sentence, our approach successfully correct this mistake. This demonstrates that cross-sentence context indeed helps resolve certain ambiguities.

\end{CJK}

\section{Related Work}

While our approach is built on top of hierarchical recurrent encoder-decoder (HRED) \cite{Sordoni2015A}, there are several key differences which reflect how we have generalized from the original model.~\newcite{Sordoni2015A} use HRED to summarize a single representation from both the current and previous sentences, which limits itself to (1) it is only applicable to encoder-decoder framework without attention model, (2) the representation can only be used to initialize decoder. In contrast, we use HRED to summarize the previous sentences alone, which provides additional cross-sentence context for NMT. Our approach is more flexible at (1) it is applicable to any encoder-decoder frameworks (e.g., with attention), (2) the cross-sentence context can be used to initialize either encoder, decoder or both.

While both our approach and \citet{Serban:2016} use Auxiliary Context mechanism for incorporating cross-sentence context, there are two main differences: 1) we have separate parameters to better control the effects of the cross- and intra-sentence contexts, while they only have one parameter matrix to manage the single representation that encodes both contexts; 2) based on the intuition that not every target word generation requires equivalent cross-sentence context, we introduce a context gate \cite{Tu:2017:TACL} to control the amount of information from it, while they don't.

At the same time, some researchers propose to use an additional set of an encoder and attention to model more information. For example, \newcite{jean2017does} use it to encode and select part of the previous source sentence for generating each target word. \newcite{calixto2017incorporating} utilize global image features extracted using a pre-trained convolutional neural network and incorporate them in NMT. As additional attention leads to more computational cost, they can only incorporate limited information such as single preceding sentence in \newcite{jean2017does}. However, our architecture is free to this limitation, thus we use multiple preceding sentences (e.g. $K=3$) in our experiments.

Our work is also related to multi-source \cite{zoph2016multi} and multi-target NMT \cite{dong2015multi}, which incorporate additional source or target languages. They investigate one-to-many or many-to-one languages translation tasks by integrating additional encoders or decoders into encoder-decoder framework, and their experiments show promising results.

\section{Conclusion and Future Work}
\label{sec:length}

% Ambiguity \& Inconsistency
We proposed two complementary approaches to integrating cross-sentence context: 1) a warm-start of encoder and decoder with global context representation, and 2) cross-sentence context serves as an auxiliary information source for updating decoder states, in which an introduced context gate plays an important role.
We quantitatively and qualitatively demonstrated that the presented model significantly outperforms a strong attention-based NMT baseline system. We release the code for these experiments at \url{https://www.github.com/tuzhaopeng/LC-NMT}.

Our models benefit from larger contexts, and would be possibly further enhanced by other document level information, such as discourse relations. We propose to study such models for full length documents with more linguistic features in future work. 
%Another interesting direction is to directly model document-level translation in a unified framework.

\section*{Acknowledgments}

This work is supported by the Science Foundation of Ireland (SFI) ADAPT project (Grant No.:13/RC/2106). The authors also wish to thank the anonymous reviewers for many helpful comments with special thanks to Henry Elder for his generous help on proofreading of this manuscript.

\balance
\bibliography{emnlp2017}
\bibliographystyle{emnlp_natbib}

\end{document}